\documentclass[sn-mathphys]{sn-jnl}

\jyear{2021}%

\usepackage{blindtext}
\usepackage{xcolor}
\usepackage{multirow}
\usepackage{graphicx,xfp}
\usepackage{booktabs,threeparttable}
\usepackage{comment}

\usepackage{caption}
\usepackage{subcaption}

\usepackage{amsmath}

\usepackage{algorithm}

\theoremstyle{thmstyleone}%
\theoremstyle{thmstyletwo}%

\theoremstyle{thmstylethree}%

\begin{document}


\title[Article Title]{Neuromorphic Spiking Neural Network Based Classification of COVID-19 Spike Sequences}



\author[1]{\fnm{Taslim} \sur{Murad}}\email{tmurad2@student.gsu.edu}

\author[1]{\fnm{Prakash} \sur{Chourasia}}\email{pchourasia1@student.gsu.edu}

\author[1]{\fnm{Sarwan} \sur{Ali}}\email{sali85@student.gsu.edu}

\author[2]{\fnm{Imdad Ullah} \sur{Khan}}\email{imdad.khan@lums.edu.pk}

\author*[1]{\fnm{Murray} \sur{Patterson}}\email{mpatterson30@gsu.edu}

\affil[1]{\orgname{Georgia State University}, \orgaddress{\street{} \city{Atlanta}, \postcode{30303}, \state{GA}, \country{USA}}}

\affil[2]{ \orgname{Lahore University of Management Sciences}, \orgaddress{\street{} \city{Lahore}, \postcode{54792}, \state{Punjab}, \country{Pakistan}}}




\abstract{
The availability of SARS-CoV-2 (severe acute respiratory syndrome coronavirus 2) virus data post-COVID has reached exponentially to an enormous magnitude, opening research doors to analyze its behavior. Various studies are conducted by researchers to gain a deeper understanding of the virus, like genomic surveillance, etc, so that efficient prevention mechanisms can be developed. However, the unstable nature of the virus (rapid mutations, multiple hosts, etc) creates challenges in designing analytical systems for it. Therefore, we propose a neural network-based (NN) mechanism to perform an efficient analysis of the SARS-CoV-2 data, as NN portrays generalized behavior upon training. Moreover, rather than using the full-length genome of the virus, we apply our method to its spike region, as this region is known to have predominant mutations and is used to attach to the host cell membrane. In this paper, we introduce a pipeline that first converts the spike protein sequences into a fixed-length numerical representation and then uses Neuromorphic Spiking Neural Network to classify those sequences. We compare the performance of our method with various baselines using real-world SARS-CoV-2 spike sequence data and show that our method is able to achieve higher predictive accuracy compared to the recent baselines.   

}







\maketitle


\section{Introduction}
\begin{figure}[h!]
  \centering
  \includegraphics[scale=0.28]{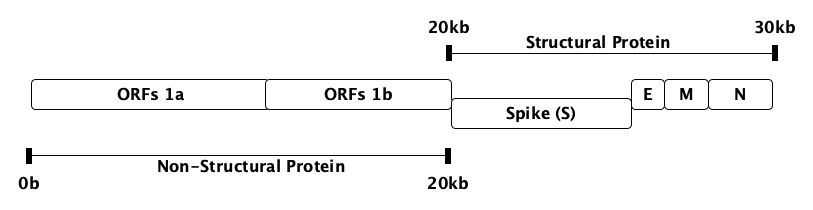}
  \caption{
  The genome of SARS-CoV-2 has a length of 30kb, and it consists of non-structural proteins (ORFs 1ab) and structural proteins (E, M, N, S). The S region is important because of its ability to attach to the host cell membrane, and also hold advantageous mutations.
  }
  \label{fig_spike_protein}
\end{figure}

The COVID-19 disease has affected millions of people across the globe~\cite{who_patient_count}. This disease is caused by the SARS-CoV-2 virus, which possesses the ability to undergo quick mutations and infect various hosts. Figure~\ref{fig_spike_protein} illustrates the genome structure of the SARS-CoV-2 virus. The length of this genome is approximately $30kb$ with spike region residing in the $21kb-25kb$ range. The spike protein region is responsible to attach to a host cell membrane, and also major mutation happens in it~\cite{kuzmin2020machine}. Thus spike region is sufficient to investigate this virus, therefore we have used only spike protein sequences in this paper. 

In computational biology, one of the crucial tasks is biological sequence classification which allows researchers to do functional analysis for building an understanding of the sequences like DNA and proteins~\cite{xia2017targeting}. Various sequence classification strategies are put forward to study the origin, structure, and behavior of the virus~\cite{ahmed2021enabling, kuzmin2020machine}. The classification models employed for this purpose include traditional machine learning (ML) approaches, such as support vector machines (SVMs)~\cite{cortes1995support}, or artificial neural networks (ANNs)~\cite{Goodfellow-et-al-2016}. 

However, the classification task has a pre-requisite of data being in a fixed-length numerical form. Therefore, several numerical embedding generation methodologies are proposed to convert biological sequences to numerical form. Some of these techniques are feature-engineering-based methods~\cite{ali2021spike2vec, NEURIPS2021_9a1de01f, kuzmin2020machine}, which contain both alignment-based and alignment-free mechanisms. However, these procedures are domain-specific and less generalized. Moreover, some other methods include signal transformation methods (Spike2Signal~\cite{ali2022spike2signal}), and image transformation methods (RP~\cite{faouzi2020pyts}, GAF~\cite{faouzi2020pyts}, MTF~\cite{faouzi2020pyts}) but they require additional data transformation steps to get the features, which can be computationally expensive. 
Furthermore, another embedding generation approach includes the usage of neural networks (NN) to get the embeddings~\cite{nguyen2021benefits}), as NNs portray more generalized behavior upon training so they could be used for heterogeneous sequences. Due to the availability of a large volume of SARS-CoV-2 data post the COVID-19 pandemic, it is more feasible to use an NN-based feature extractor. However, their need for extensive training with a large dataset to achieve good performance is again a computational overhead. 

Recently, spiking neural networks (SNNs)~\cite{maass1997networks, ponulak2011introduction, guerguiev2017towards} have emerged as a numerical feature extractor and classifier for biological sequences. SNNs are artificial neural networks that more closely mimic natural neural networks. Unlike ANNs whose neurons exhibit non-linear behavior with being continuous function approximators following a common clock cycle to operate, neurons of SNNs use asynchronous spikes to signal the occurrence of some characteristic event by digital and temporally precise action potentials~\cite{pfeiffer2018deep}. In this paper, we propose an SNN-based method to classify the spike protein sequences of the SARS-CoV-2 virus. Given any spike sequence, our method follows an alignment-free pipeline of extracting numerical features from the sequence and using those features to perform classification.   

In this paper, our contributions are the following: 
\begin{enumerate}
    \item We propose an alignment-free end-to-end classification pipeline for biological sequences using spiking neural networks (SNN), which first converts the sequences into numerical form and using these numerical features perform Classification.
    \item Compared to existing neural network-based baselines, we show that SNN is comparatively more stable in terms of predictive accuracy for biological sequence analysis.
    \item We show from the experimental results that we can achieve higher predictive accuracy by using the spike region of the protein sequence only.
\end{enumerate}

Our work is distributed in the manuscript as follows: Section~\ref{sec_related_work} deals with talking about related work, Section~\ref{sec_proposed_approach} highlights our proposed method, Section~\ref{sec_experimental_setup} discusses the details of experiments, Section~\ref{sec_results_discussion} talks about the results, and Section~\ref{sec_conclusion} concludes the paper.

\section{Related Work}\label{sec_related_work}
Many works are done in the domain of biological sequence classification. Some researchers have explored traditional ML models, like SVM, for doing this task. The ML-based classification requires mapping the sequences to numerical form first. Feature-engineering-based embedding generation methods are popular in this regard, like n-gram-based vector representation~\cite{ali2021k, ali2021effective}, Sparse~\cite{hirst1992prediction}, Spike2Vec~\cite{ali2021spike2vec}, PWM2Vec~\cite{ali2022pwm2vec} etc. However, usually, these techniques require sequence alignment, which is a computationally expensive operation. Additionally, feature-engineering-based methods are domain-specific and they may not be able to generalize to heterogeneous (different types) data. 

Moreover, ANNs are also gaining popularity for performing biological sequence classification because of their generalizability property and larger sequence data's availability, like protein classification~\cite{zainuddin2008radial}, DNA binding site prediction~\cite{zhou2016cnnsite} etc. Some of the broader categories of ANN-based classification, depending on the feature extractor method, can be neural networks-based, image transformation-based, and signal transformation based. In neural network-based methods, they employ an ANN model to extract the sequence's features and then perform classification of those features. Like, ~\cite{nguyen2021benefits}) uses an auto-encoder to do sequence classification. This network consists of symmetric encoder and decoder parts, and the output for the encoder corresponds to the feature vectors of the sequence. However, they possess a computational overhead of training with a large dataset for achieving good performance. 

Furthermore, the image transformation-based approaches e.g. RP, GAF, MTF~\cite{faouzi2020pyts}, uses a mechanism to transform the sequence to an image first, and then use this graphical form to perform classification using ANN models. Likewise, a signal transformation-based approach, like Spike2Signal~\cite{ali2022spike2signal}, maps a protein sequence to signal-like numerical data by assigning an integer value to each amino acid of the sequence and then uses this signal-like data for classification. It can be noted that both image transformation and signal transformation approaches have an added step of data transformation which is a computationally expensive step. 

Spiking neural networks-based sequence classifications are also designed by some of the previous works~\cite{ponulak2011introduction, zhou2018deep} and they illustrate that SNNs can achieve competitive performance compared to traditional machine learning methods while being more efficient and requiring less data~\cite{hunsberger2016training}. However, these previous studies have mostly focused on simple biological sequences, such as DNA sequences with a fixed length. In this work, we propose the use of SNNs for classifying more complex biological sequences, such as proteins with varying lengths. To the best of our knowledge, this is the first study that investigates the use of SNNs for this task.

\section{Proposed Approach}\label{sec_proposed_approach}
This section highlights our proposed end-to-end alignment-free method to do the classification of SARS-CoV-2 protein spike sequences using a neuromorphic design based on the spiking neural network (SNN). The neuromorphic design field, which has the purpose of building machines that mimic the structure of the brain, has recently taken the realm of machine learning into account. Although the application of neuromorphic architecture to traditional tasks, such as image recognition and logistic regression, is associated with many challenges, the advancements in ML have opened new breakthroughs in this area. Furthermore, the replacement of DL model neurons with spiking neurons can cause neuromorphic computing to improve the efficiency and performance of predictions. 

In SNN, based on the exceeded threshold value the neurons emit spikes or electrical impulses in response to the input. In case of input being lower than the threshold, the pre-activation value gradually decreases. This phenomenon is equivalent to a time-dependent version of the ReLU activation function in which at different time steps there is either a spike or no spike. Due to the similarity between the brain's information accumulation and release procedure to SNN, they are considered more biologically realistic than traditional ML and DL models. 
For input data $X$, after multiplying it with a weight matrix $W$ the result is passed on to a decayed version of the information inside the neuron. This information exists from the previous time step/time tick ($\Delta t$ time elapsed). To serve the aim of gradually reducing the inner activation a $decay\_multiplier$ is used, which prevents the accumulation of stimuli for a long time on a neuron. We use a $decay\_multiplier$ of $0.9$ value in our model. After that, the neuron's activation function is computed based on a threshold. The neuron's inner state is reset on its fire by subtracting the activation from its inner state. This will stop the neuron from firing constantly upon being activated once and also it will isolate each firing event from the other by clipping the gradient through time. Finally, in the classification layer, the spiking neurons values are averaged over the time axis and plugged to the softmax cross-entropy loss function for back-propagation. 

Moreover, data processing in SNN contains a temporal dimension and this dimension is incorporated into Artificial Neural Networks (ANNs) by allowing the signal to accumulate over time in a pre-activation phase. We use the SNN for doing classification of SARS-CoV-2 spike sequences.

Our system can be applied to data with varying lengths. We convert the input data to its corresponding one-hot encoding (OHE) vector and pass this OHE-based vector to our model. A zero-padding technique is employed on the OHE vector to handle the sequences of varying lengths. To train our system for experiments, we use cross-entropy loss function, ADAM optimizer, $100$ epochs, and $0.001$ learning rate. Our model consists of two linear layers.

\section{Experimental Evaluation}\label{sec_experimental_setup}
This section discusses the dataset used to perform the experiments, along with all the evaluation metrics. It also highlights the baseline models employed for performance comparison. We follow the $70\%-30\%$ train-test data split with k-folds ($k=5$) to conduct the experiments. All experiments are executed using an Intel(R) Xeon(R) CPU E7-4850 v4 @ 2.10GHz having Ubuntu 64-bit OS (16.04.7 LTS Xenial Xerus) with 3023 GB memory. 

\subsection{Dataset Statistics}
We employ a dataset about spike sequences of the SARS-CoV-2 virus taken from GISAID~\cite{gisaid_website_url} to investigate the performance. This dataset has $7000$ sequences and is referred to as the Spike7k dataset. We have randomly selected a subset of $7k$ sequences from the whole GISAID dataset with the aim of mimicking real-world scenarios and preserving the original lineage distribution of the coronavirus to avoid any biases in our results. Some of the common lineages are given names by the world health organization, like B.1.1.7 is Alpha, AY.12 is Delta, B.1.429 is Epsilon, AY.4 is Delta, etc. These $7000$ sequences encode details of the $22$ Lineage of coronavirus. The detailed distribution is illustrated in Table~\ref{tbl_dataset_statistics_spike7k}.

\begin{table}[h!]
    \centering
     \resizebox{0.45\textwidth}{!}{
    \begin{tabular}{lc|lc}
    \toprule
        Lineage & Frequency & Lineage & Frequency \\
        \midrule 
        B.1.1.7 & 3369 & R.1 & 32 \\
        B.1.617.2 & 875 & AY.4 & 593  \\
        B.1.2 & 333 & B.1 & 292 \\
        B.1.177 & 243 & P.1 & 194  \\
        B.1.1 & 163 & B.1.429 & 107  \\
        B.1.526 & 104 & AY.12 & 101  \\
        B.1.160 & 92 & B.1.351 & 81  \\
        B.1.427 & 65 & B.1.1.214 & 64  \\
        B.1.1.519 & 56 & D.2 & 55  \\
        B.1.221 & 52 & B.1.177.21 & 47  \\
        B.1.258 & 46 & B.1.243 & 36  \\
        \bottomrule
    \end{tabular}
    }
    \caption{ The distribution of $7k$ sequences across all the Lineages in the Spike7k dataset is shown in this table.}
    \label{tbl_dataset_statistics_spike7k}
\end{table}

\subsection{Evaluation Metrics and Classifiers}
To measure the performance of our proposed and baseline models we utilize the accuracy, precision, recall, F1 (weighted), F1 (macro), Receiver Operator Characteristic Curve Area Under the Curve (ROC AUC), and training runtime. The reported value for each evaluation metric is an average value over five runs. The ROC AUC is computed using the one-vs-rest method. 

The machine learning classifiers used to get the evaluation metrics of the neural network baseline are Support Vector Machine (SVM), Naive Bayes (NB), Multi-Layer Perceptron (MLP), K-Nearest Neighbors (KNN), Random Forest (RF), Logistic Regression (LR), and Decision Tree (DT) classifiers. 

The image transformation baselines are evaluated using 3-Layer CNN, 4-Layer CNN, and RESNET34 deep learning models. The 3 Layer CNN and 4 Layer CNN correspond to neural networks with 3 and 4 convolution layers respectively. The RESNET34 refers to a pre-trained RESNET34~\cite{he2016deep} model. All these DL models are trained using ADAM optimizer and negative log-likelihood loss function. 

The evaluation of signal transformation baseline is done using Fully Convolution Network (FCN)~\cite{wang2017time}, LSTM 3 Layer Bidirect~\cite{hochreiter1997long}, and mWDN networks~\cite{wang2018multilevel}.

\subsection{Baseline Methods}
We select the baseline methods from 3 different domains to investigate the performance and these domains are:
\begin{enumerate}
    \item An end-to-end classification pipeline using a neural network, which generates feature embeddings from the raw input and performs classification based on the generated features.
    \item Sophisticated image classifiers that take transformed images and classify them.
    \item Time sequence classifiers, which first transform sequence into signal-like data and then apply signal NN for classification.
\end{enumerate}
The details of each baseline are as follows:

\subsubsection{Autoencoder + Neural Tangent Kernel~\cite{nguyen2021benefits}}
This approach generates a low-dimensional embedding of protein sequences through an encoder neural network. The architecture of the encoder consists of a stack of dense layers with LeakyReLU activation function, batch normalization, and dropout. It's accompanied by a symmetric decoder for training and a reconstruction loss is used to optimize the whole network. Then the low-dimensional embeddings are utilized to compute Neural Tangent Kernel (NTK). The NTK is a kernel function that measures the similarity between two input sequences based on the geometry of the neural network's decision boundary. This method is efficient for large and high-dimensional data. We have used a $4$-layer autoencoder with ADAM optimizer, MSE loss function, and $100$ epochs to train this model for our experiments.

\subsubsection{Image Transformation}
As image-based classifiers are considered state-of-the-art for the classification tasks, therefore we wanted to draw a performance comparison of these classifiers with our proposed model for protein sequence classification which is why we used image transformation methods as baselines.

Image transformation represents a set of methods that transform signal-based sequences into images to perform analysis. The sequences are transformed into signals following the Spike2Signal~\cite{ali2022spike2signal} method. The three models belonging to the category of image transformation reported in the experiments are as follow,

\paragraph{Recurrent Plot (RP)} RP~\cite{faouzi2020pyts} is used to get a 4D image of $360$x$360$ size corresponding to a signal-based represented spike sequence. The 4D refers to three color channels and one alpha channel. This image illustrates the distance between the trajectories. Once the trajectories are extracted from the signal, the pair-wise distance between them is computed to get a graphical form. For a given signal $x_i$, it's trajectories with $m$ dimensions and $\tau$ time-delay are defined as:
\begin{equation}
    \vec{x}_i = (x_i, x_{i + \tau}, \ldots, x_{i + (m - 1)\tau}), \quad \forall i \in \{1, \ldots, n - (m - 1)\tau \} 
\end{equation}

\paragraph{Gramian Angular Field (GAF)} Given a pair of signal data, GAF~\cite{faouzi2020pyts} extracts their temporal correlation matrix using the formula,
\begin{equation}
    \begin{aligned}
    \tilde{x}_i = a + (b - a) \times \frac{x_i - \min(x)}{\max(x) - \min(x)}, \quad \forall i \in \{1, \ldots, n\}  \\
    \phi_i = \arccos(\tilde{x}_i), \quad \forall i \in \{1, \ldots, n\}  \\
    GAF_{i, j} = \cos(\phi_i + \phi_j), \quad \forall i, j \in \{1, \ldots, n\}  \\
    \end{aligned}
\end{equation}
\noindent where $a < b$ and both fall in the range $[-1,1]$ and represents the range to rescale the original signal. $\phi_i$ shows the polar coordinates of the scaled signal.

\paragraph{Markov Transition Field (MTF)} The sequence data is discretized into Q quantile bins by MTF~\cite{faouzi2020pyts}. Then a square matrix $W$ with dimensions $Q$ is created from each bin. After that, from the matrix $W$ having transition probability from $q_i$ to $q_j$ amino acids, a matrix $M$ is computed. This $M$ is used for visualization. 

\begin{figure}[h!]
 \centering
    \begin{subfigure}{.33\textwidth}
        \centering
        \includegraphics[scale = 0.3] {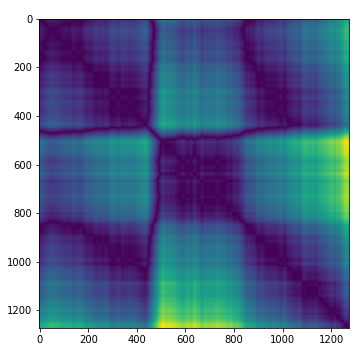}
        \caption{RP}
        \label{fig_a}
    \end{subfigure}%
    \begin{subfigure}{.33\textwidth}
        \centering
        \includegraphics[scale = 0.3] {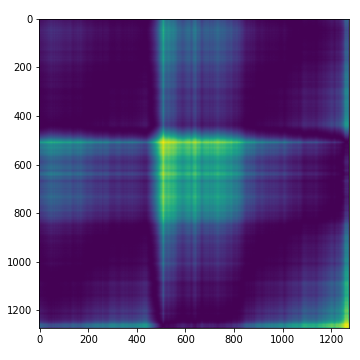}
        \caption{GAF}
        \label{fig_b}
    \end{subfigure}%
    \begin{subfigure}{.33\textwidth}
        \centering
        \includegraphics[scale = 0.3] {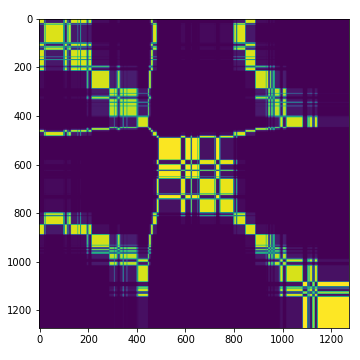}
        \caption{MTF}
        \label{fig_c}
    \end{subfigure}%
    \\
    \begin{subfigure}{.33\textwidth}
        \centering
        \includegraphics[scale = 0.3] {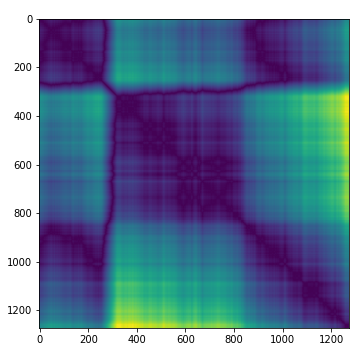}
        \caption{RP}
        \label{fig_d}
    \end{subfigure}%
    \begin{subfigure}{.33\textwidth}
        \centering
        \includegraphics[scale = 0.3] {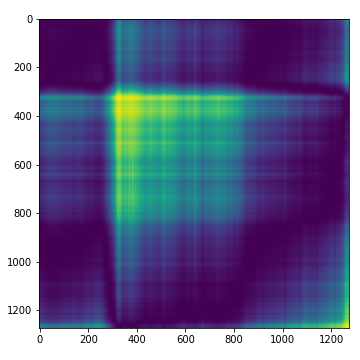}
        \caption{GAF}
        \label{fig_e}
    \end{subfigure}%
    \begin{subfigure}{.33\textwidth}
        \centering
        \includegraphics[scale = 0.3] {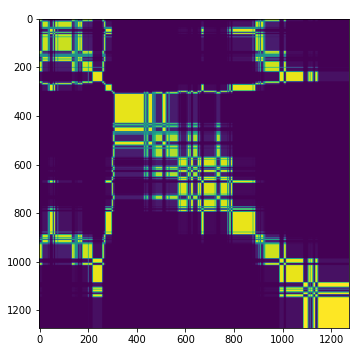}
        \caption{MTF}
        \label{fig_f}
    \end{subfigure}%
    \caption{ The images from ($a-c$) are against a spike sequence corresponding to Lineage AY.12, while ($d-f$) show images of a sequence from B.1.526 Lineage. They are created using RP, GAF, and MTF approaches.}
    \label{images_overview}
\end{figure}

Some of the examples of images generated from RP, GAF, and MFT image transformation techniques are illustrated in Figure~\ref{images_overview}. It shows a set of images for Lineage AY.12 and B.1.526 from our dataset. For a sequence belonging to a Lineage, we can observe that the image patterns differ corresponding to each transformation technique, which shows that the information is differently captured within the image depending on the underlying transformation technique used, and this can be beneficial for classification.

\subsubsection{Signal Transformation}
Due to the presence of some NN models specifically designed for signal-based data, we used signal transformation as a baseline because we wanted to explore those NN models' performance on protein sequence classification.
The signal transformation includes the Spike2Signal~\cite{ali2022spike2signal} approach which is used to map spike sequences into signal-like numerical representations to make them compatible with deep learning models. This method assigns integer values to amino acids in the spike sequences.

\section{Results and Discussion}\label{sec_results_discussion}
In this section, we discuss the classification results of our proposed system and compare the results with the baseline methods. Table~\ref{tbl_results} summarizes the results of the neural network, image transformation, signal transformation, and spiking neural network (our proposed one) based approaches. 

The results illustrate that our method has drastically improved the performance for all the evaluation metrics as compared to the neural network-based baseline. This behavior is may be caused by the underlying feature extractor of the baseline being not able to produce optimal features which degrade the classification performance. However, our system uses an advanced SNN model to generate optimal features which can deliver good classification performance.

Similarly, we can view a clear performance improvement for all the metrics using our model over the signal transformation baseline. The low classification performance of the signal transformation baseline can also be associated with using a sub-optimal feature generation mechanism. 

Moreover, we can observe that SNN is performing better in terms of accuracy, precision, and F1 weighted score as compared to the image transformation baseline. However, SNN is not optimized for recall, F1 macro, and ROC AUC score but it yields comparable results for these metrics. Since the image transformation procedures are task-specific, SSNs can portray more generalized behavior after being trained, therefore overall SNNs possess good performance ability among the all mentioned baselines. The good performance of SNN also indicates that it is able to handle class imbalance efficiently.

\begin{table}[h!]
    \centering
    \resizebox{0.99\textwidth}{!}{
    \begin{tabular}{cp{2.5cm}p{2.5cm}p{0.9cm}p{0.9cm}p{0.9cm}p{1.1cm}p{1.1cm}p{1.1cm}|p{1.8cm}}
    \toprule
        Approach & Embedding & Algo. & Acc. & Prec. & Recall & F1 (Weig.) & F1 (Macro) & ROC AUC & Train Time \\
        \midrule
        
          \multirow{7}{1.2cm}{Neural Network} & \multirow{7}{1.2cm}{Autoencoder + NTK} & SVM & 0.480 & 0.503 & 0.480 & 0.478 & 0.146 & 0.684 & 0.011 Sec. \\
 & & NB & 0.507 & 0.464 & 0.507 & 0.458 & 0.224 & 0.653 & 0.002 Sec. \\
&  & MLP & 0.467 & 0.460 & 0.467 & 0.449 & 0.194 & 0.653 & 0.917 Sec. \\
&  & KNN & 0.413 & 0.337 & 0.413 & 0.363 & 0.083 & 0.581 & 0.002 Sec. \\
&  & RF & 0.520 & 0.490 & 0.520 & 0.487 & 0.199 & 0.687 & 0.185 Sec. \\
&  & LR & 0.507 & 0.484 & 0.507 & 0.482 & 0.175 & 0.653 & 0.009 Sec. \\
&  & DT & 0.520 & 0.546 & 0.520 & 0.525 & 0.189 & 0.694 & \textbf{0.001} Sec. \\
\midrule

         \multirow{16}{2.5cm}{Image Transformation} & 
        \multirow{2}{2.5cm}{Recurrent Plot}
        & \multirow{4}{1.9cm}{3 Layer CNN} & \multirow{2}{*}{0.780} & \multirow{2}{*}{0.737} & \multirow{2}{*}{\textbf{0.777}} & \multirow{2}{*}{0.746} & \multirow{2}{*}{0.398} & \multirow{2}{*}{0.713} & \multirow{2}{*}{1.62 Hours} \\
        &&&&&&&&& \\ 
        \cline{4-10}
        & 
        \multirow{1}{2.5cm}{Gramian Angular Field}
        & & \multirow{2}{*}{0.750} & \multirow{2}{*}{0.713} & \multirow{2}{*}{0.754} & \multirow{2}{*}{0.717} & \multirow{2}{*}{0.411} & \multirow{2}{*}{0.706} & \multirow{2}{*}{1.58 Hours} \\
        &&&&&&&&& \\ 
        \cline{2-10}
        
       & 
        \multirow{2}{2.5cm}{Recurrent Plot}
        & \multirow{4}{1.9cm}{4 Layer CNN} & \multirow{2}{*}{0.780} & \multirow{2}{*}{0.750} & \multirow{2}{*}{\textbf{0.777}} & \multirow{2}{*}{0.757} & \multirow{2}{*}{\textbf{0.468}} & \multirow{2}{*}{\textbf{0.752}} & \multirow{2}{*}{1.7 Hours} \\
        &&&&&&&&& \\ 
        \cline{4-10}
        & 
        \multirow{1}{2.5cm}{Gramian Angular Field}
        & & \multirow{2}{*}{0.764} & \multirow{2}{*}{0.718} & \multirow{2}{*}{0.764} & \multirow{2}{*}{0.732} & \multirow{2}{*}{0.417} & \multirow{2}{*}{0.716} & \multirow{2}{*}{1.69 Hours} \\
        &&&&&&&&& \\ 
        \cline{2-10}

         & 
        \multirow{2}{2.5cm}{Recurrent Plot}
        & \multirow{4}{1.9cm}{RESNET34} & \multirow{2}{*}{0.599} & \multirow{2}{*}{0.587} & \multirow{2}{*}{0.599} & \multirow{2}{*}{0.553} & \multirow{2}{*}{0.172} & \multirow{2}{*}{0.609} & \multirow{2}{*}{12.3 Hours} \\
        &&&&&&&&& \\ 
        \cline{4-10}
        & 
        \multirow{1}{2.5cm}{Gramian Angular Field}
        & & \multirow{2}{*}{0.712} & \multirow{2}{*}{0.669} & \multirow{2}{*}{0.712} & \multirow{2}{*}{0.685} & \multirow{2}{*}{0.342} & \multirow{2}{*}{0.674} & \multirow{2}{*}{11.4 Hours} \\
        &&&&&&&&& \\ 
        
         \midrule
         \multirow{6}{2.5cm}{Signal Transformation} & 
        \multirow{6}{1.9cm}{Numerical Representation}
        & \multirow{2}{0.8cm}{mWDN} & \multirow{2}{*}{0.527} & \multirow{2}{*}{0.348} & \multirow{2}{*}{0.527} & \multirow{2}{*}{0.394} & \multirow{2}{*}{0.078} & \multirow{2}{*}{0.523} & \multirow{2}{*}{11 Hours} \\
        &&&&&&&&& \\
        & & \multirow{2}{0.8cm}{FCN} & \multirow{2}{*}{0.615} & \multirow{2}{*}{0.566} & \multirow{2}{*}{0.615} & \multirow{2}{*}{0.568} & \multirow{2}{*}{0.134} & \multirow{2}{*}{0.563} & \multirow{2}{*}{1.1 Hours} \\
        &&&&&&&&& \\ 
        & & \multirow{2}{2.7cm}{LSTM 3 Layer Bidirect.} & \multirow{2}{*}{0.740} & \multirow{2}{*}{0.701} & \multirow{2}{*}{0.740} & \multirow{2}{*}{0.709} & \multirow{2}{*}{0.377} & \multirow{2}{*}{0.682} & \multirow{2}{*}{22 Hours} \\
        &&&&&&&&& \\ 
        \cmidrule{1-10}
        \multirow{2}{2.7cm}{Spiking Neural Network (ours)} & \multirow{2}{1.9cm}{Numerical Representation} & \multirow{2}{*}{-} & \multirow{2}{*}{\textbf{0.810}} & \multirow{2}{*}{\textbf{0.790}} & \multirow{2}{*}{0.710} & \multirow{2}{*}{\textbf{0.782}} & \multirow{2}{*}{0.426} & \multirow{2}{*}{0.717} & \multirow{2}{*}{3 days} \\
        &&&&&&&&& \\

         \bottomrule
         \end{tabular}
    }
    \caption{Classification results for different methods on SARS-CoV-2 data. The best values are shown in bold.}
    \label{tbl_results}
\end{table}

\section{Conclusion}\label{sec_conclusion}
In this paper, we propose using spiking neural networks (SNNs) for biological sequence classification. We describe our proposed method, which uses SNNs to classify proteins with varying lengths, and evaluate its performance on a benchmark dataset. Our results show that our proposed method achieves competitive performance compared to state-of-the-art methods while being more efficient and requiring less data.

Overall, our work suggests that SNNs are a promising approach to biological sequence classification, and they can potentially improve our understanding of biological sequences and their functions. Further research is needed to explore the full potential of SNNs in this domain and to develop new methods and techniques for improving their performance. In the future, we can explore the scalability and  robustness of the SNNs in terms of protein sequence classification. Applying SNN on nucleotide sequence and evaluating its performance is also an interesting future direction.









\bibliography{references}

\end{document}